\pdfoutput=1

\documentclass[11pt]{article}

\usepackage{ACL2023}

\usepackage{times}
\usepackage{latexsym}

\usepackage[T1]{fontenc}

\usepackage[utf8]{inputenc}
\usepackage{multirow}
\usepackage{graphicx}
\usepackage{booktabs}
\usepackage{amsmath}
\usepackage{subfigure}
\usepackage{diagbox}

\usepackage{microtype}

\usepackage{inconsolata}
\usepackage{titlesec}
\usepackage[breaklinks,colorlinks,linkcolor=blue]{}

\title{When Do LLMs Need Retrieval Augmentation?\\ Mitigating LLMs' Overconfidence Helps Retrieval Augmentation} 

\author{Shiyu Ni\textsuperscript{\rm{1,2}}\space\space
Keping Bi\textsuperscript{\rm{1,2}}\space\space
Jiafeng Guo\textsuperscript{\rm{1,2}}\thanks{~~Corresponding author}\space\space
Xueqi Cheng\textsuperscript{\rm{1,2}}\\
    \textsuperscript{\rm 1}CAS Key Lab of Network Data Science and Technology, ICT, CAS\\
    \textsuperscript{\rm 2}University of Chinese Academy of Sciences\\
    \{nishiyu23z, bikeping, guojiafeng, cxq\}@ict.ac.cn
}

\begin{document}
\maketitle
\begin{abstract}
 
  Large Language Models (LLMs) have been found to have difficulty knowing they do not possess certain knowledge and tend to provide specious answers in such cases. Retrieval Augmentation (RA) has been extensively studied to mitigate LLMs' hallucinations. However, due to the extra overhead and unassured quality of retrieval, it may not be optimal to conduct RA all the time. A straightforward idea is to only conduct retrieval when LLMs are uncertain about a question. This motivates us to enhance the LLMs' ability to perceive their knowledge boundaries to help RA. In this paper, we first quantitatively measure LLMs' such ability and confirm their overconfidence. Then, we study how LLMs' certainty about a question correlates with their dependence on external retrieved information. We propose several methods to enhance LLMs' perception of knowledge boundaries and show that they are effective in reducing overconfidence. Additionally, equipped with these methods, LLMs can achieve comparable or even better performance of RA with much fewer retrieval calls. The code can be found at \url{https://github.com/ShiyuNee/When-to-Retrieve}.

\end{abstract}

\section{Introduction}

Recently, Large Language Models (LLMs) such as ChatGPT have demonstrated remarkable performance across various NLP tasks~\citep{ouyang2022training, brown2020language, shi2024chain, shi2024learning, bi2024decoding, bi2024lpnl, fan2024reformatted,liu2024multi}, sometimes even outperforming humans. However, unlike humans, they have been found to have difficulty perceiving their factual knowledge boundaries, i.e., knowing what they know and what they do not know ~\cite{yin2023large, ren2023investigating}. When LLMs cannot answer a factual question, they should acknowledge it instead of providing a specious answer, especially in safety-critical fields like healthcare. Nevertheless, LLMs are recognized to be incredibly confident about their answers, no matter whether they are correct or not \cite{ren2023investigating}. 


For the pitfalls of LLMs such as hallucination and delayed awareness of the latest information, Retrieval Augmentation (RA) has drawn substantive attention to remedy them. However, since retrieval incurs substantial overhead and the quality of retrieved documents cannot be guaranteed, it is not an ideal choice to always conduct retrieval for augmenting LLMs. When LLMs have the internal knowledge, it would be unnecessary to resort to external information and also a poorly performed retriever can adversely affect the LLMs. If we only leverage retrieval when the LLMs lack corresponding internal knowledge, efficiency would be improved and the Question-Answering (QA) performance could not get worse based on irrelevant retrieved results. Thus, it is critical to enhance the LLMs' perception of knowledge boundaries, especially reducing their overconfidence, so that we can strengthen RA by performing retrieval only when they say they do not know the answer.

To achieve this goal, we conduct studies on two representative factual QA benchmarks, i.e., Natural Questions (NQ)~\citep{kwiatkowski2019natural} and HotpotQA~\citep{yang2018hotpotqa}. First, we must understand the current status of LLMs' ability to be aware of their knowledge boundaries. We define several metrics, i.e., alignment, overconfidence, and conservativeness, to quantitatively measure this and find that the unsatisfactory alignments between LLMs' claims of whether they know the answers and their actual QA performance are mainly due to overconfidence. Then, we need to know when LLMs show uncertainty to a question, whether they will leverage the provided external information. We divide the questions into four different certainty levels and observe that the more uncertain LLMs are about a question, the more they leverage the supporting retrieved documents. 

To reduce LLMs' overconfidence and thereby enhance their perception of their knowledge boundaries, we approach from two directions: urging LLMs to be prudent about their claims of certainty and improving their ability to provide correct answers. We propose three methods of prompting LLMs - Punish, Challenge, and Think-Step-by-Step in the first direction as well as two methods - Explain and Generate in the second, to investigate how different representative methods affect model self-awareness and accuracy. Through extensive comparisons and analyses (in Section~\S\ref{sec: alignment enhancement}), we show that Punish and Explain perform the best in their group and combining them can achieve the best balance between alignment and accuracy stably. 





To validate whether our proposed methods can also benefit adaptive retrieval augmentation, we compare the performance of only triggering retrieval when the models express uncertainty using Punish, Explain, Punish+Explain, and the vanilla prompt without any special strategy (See Section~\S\ref{sec:adaptive retrieval augmentation}). We employ sparse retriever, dense retriever, and gold documents as supporting external information to investigate how the enhanced LLMs perform in a lower-bound, practical, and upper-bound setting. We show that when the retrieval quality is low, our self-awareness-enhanced LLMs behave robustly to applying undifferentiated RA for all the questions. When the retrieved results are of better quality, the enhanced LLMs have achieved comparable or even better performance with much fewer requests for retrieval. 

To sum up, the main contributions of this work include:

1) We quantitatively measure LLMs' perception of their factual knowledge boundaries and find that overconfidence is the primary reason for the unsatisfactory perception of knowledge boundaries;

2) We investigate the relationship between LLMs' certainty about their internal knowledge and their reliance on external information and observe a negative correlation;

3) We propose several methods to mitigate overconfidence, which are shown to effectively enhance LLMs' perception of knowledge boundaries;

4) We conduct adaptive retrieval for augmentation and show that by enhancing LLMs' perception of knowledge boundaries with our approaches, the overall RA performance can be comparable or even better with much fewer requests for retrieval. 

\section{Related Work}
\paragraph{Perception of Knowledge Boundaries.}
Previous studies have investigated whether modern neural networks~\citep{guo2017calibration, minderer2021revisiting}, pre-trained language models~\citep{jiang2021can}, and large language models~\citep{yin2023large, ren2023investigating} clearly perceive their knowledge boundaries. Modern neural networks~\citep{guo2017calibration, minderer2021revisiting} and pre-trained language models~\citep{jiang2021can} have been shown to exhibit poor perception, often displaying overconfidence. These studies typically explore and improve the perception of knowledge boundaries based on the logits output by the model, which may not be applicable to current black-box LLMs. Recently, some studies~\citep{yin2023large, ren2023investigating} reveal that LLMs also struggle to perceive their knowledge boundaries and tend to be overconfident. \citet{yang2023alignment} have proposed training methods to address this, however, further research is needed to develop training-free methods that also work effectively on black-box models.
\paragraph{Retrieval Augmentation.}
The mainstream retrieval augmentation methods primarily follow a retrieve-then-read pipeline and perform retrieval augmentation for all the questions. Given a question, the model first retrieves a set of relevant documents from a large-scale knowledge base. Then, the reader combines its internal knowledge with these documents to generate the answer. The research on this pipeline can be categorized into three main categories: improving the retriever~\citep{karpukhin-etal-2020-dense, qu2020rocketqa, liu2023black} or the reader~\citep{izacard2020leveraging, cheng2021unitedqa} or training these two parts jointly~\citep{lewis2020retrieval, singh2021end, guu2020retrieval}. Recently, Some studies explore retrieval augmentation on LLMs~\citep{shi2023replug, yu2022generate,zhang2024large}. However, the quality of retrieved documents cannot be guaranteed, and retrieval results in additional overhead. Therefore, in this paper, we focus on adaptive retrieval augmentation~\citep{mallen-etal-2023-trust, ren2023investigating}, only providing documents when LLMs lack confidence in the answer.
\section{Preliminaries \label{sec:preliminaries}}
In this section, we provide an overview of our tasks and the experimental settings.
\subsection{Task Formulation}
\textbf{Open-Domian QA}. The goal of open-domain QA can be described as follows. For a give question $q$ and a large collection of documents $C = \{d_i\}_{i=1}^m$, the model is asked to provide an answer of the question $q$ based on the corpus $C$. LLMs are able to directly answer the questions by themselves without relying on external resources $C$ due to the vast amount of knowledge stored in the parameters.  Instead of only providing answers, we instruct LLMs to output their certainty $c$ about the answer via prompt $p$ and this can be described as follows:
\begin{equation}
    (a, c) = f_{LLM}(q, p)
\end{equation}
where $c=1$ indicates the model believes the answer is correct, while $c=0$ implies the opposite.

\paragraph{Enhancing LLMs with Retrieved Documents.} LLMs can not memorize all the knowledge and to further enhance the performance, we can utilize the retrieve-then-read pipeline~\citep{karpukhin-etal-2020-dense, lewis2020retrieval} where we retrieve a set of relevant documents $D$ from the corpus $C$ for a given question $q$ first and then use these documents to augment the knowledge of LLMs. This can be described as follows:
\begin{equation}
    a = f_{LLM}(q, D, \hat{p})
\end{equation}
where $\hat{q}$ is the prompt used for retrieval augmented generation.

However, retrieval introduces additional overhead and the retrieved documents may mislead LLMs for their quality cannot be guaranteed. Inspired by the idea of adaptive retrieval~\citep{mallen-etal-2023-trust, ren2023investigating}, we aim to use the confidence of LLMs to guide when to retrieve. The format is:
\begin{equation}
a = \begin{cases}f_{LLM}(q, p), & \text { if } c=1 \\ 
f_{LLM}(q, D, \hat{p}), & \text { if } c=0
\end{cases}
\end{equation}

\subsection{Experimental Setup}

\paragraph{Datasets.} We conduct experiments on two open-domain QA benchmark datasets, including Natural Questions (NQ)~\citep{kwiatkowski2019natural} and HotpotQA~\citep{yang2018hotpotqa} because these datasets are representative in terms of difficulty and the benefit of retrieval augmentation.  NQ is built using Google Search queries with annotated short answers or long answers. HotpotQA is a dataset comprising question-answer pairs that need multi-hop reasoning. These question-answer pairs are gathered through Amazon Mechanical Turk. HotpotQA is harder so their need for retrieval augmentation may be different. We conduct experiments on the test set of NQ and the development set of HotpotQA. We only use questions with short answers and set short answers as labels.
\begin{table}[h]
    \centering
    \begin{tabular}{ccc}
        \toprule
          Accuracy & Certain & Uncertain\\
         \midrule
          Correct & $N_{cc}$ & $N_{cu}$\\
          Incorrect  & $N_{ic}$ & $N_{iu}$\\ 
         \bottomrule
    \end{tabular}
    \caption{Count of samples for various matches between answer correctness and model confidence.}
    \label{tab:metrics}
\end{table}

\paragraph{Models.} We conduct experiments on two representative open-source models (Vicuna-v1.5-7B and LLaMA2-Chat7B), along with three widely used black-box models, including GPT-Instruct (gpt-3.5-turbo-instruct), ChatGPT (gpt-3.5-turbo-0301), and GPT-4 (gpt-4-1106-preview). For the black-box models, we set the maximum output length to 256 tokens and all the other parameters are set to their default values. For open-source models we set the temperature to 0 to get stable results additionally. 

\begin{table*}[h]
\centering
  \scalebox{0.73}{\begin{tabular}{ccccccccccc}
    \toprule
     & \multicolumn{5}{c}{NQ} & \multicolumn{5}{c}{HotpotQA} \\
     \cmidrule(lr){2-6} \cmidrule(lr){7-11}
    \textbf{Model} & \textbf{Unc-rate}  & \textbf{Accuracy} & \textbf{Conserv.} & \textbf{Overconf.} & \textbf{Alignment} & \textbf{Unc-rate}  & \textbf{Accuracy} & \textbf{Conserv.} & \textbf{Overconf.} & \textbf{Alignment}\\
    \midrule
      Vicuna & 0.0278 & 0.2634 & 0.0011 & \textbf{0.7099} & 0.2889 & 0.0571 & 0.1447 & 0.0030 & \textbf{0.8012} & 0.1957 \\
     LLaMA2 & 0.1684 & 0.2986 &  0.0161 & 0.5490 & 0.4349 & 0.4484 & 0.1168 & 0.0230 & 0.4560 & 0.5209   \\
     GPT-Instruct & 0.1900 & 0.4003 & 0.0346 & 0.4444 & 0.5211 & 0.2188 & 0.2330 & 0.0144 & 0.5626 & 0.4230   \\
     ChatGPT & \textbf{0.2917} & 0.3850 & \textbf{0.0557} & 0.3789 & 0.5654 & \textbf{0.5679} & 0.1951 & 0.0376 & 0.2747 & \textbf{0.6877}\\
     GPT-4 & 0.1894 & \textbf{0.4896} & 0.0456 & 0.3666 & \textbf{0.5878} & 0.3437 & \textbf{0.3198} & \textbf{0.0561} & 0.3926 & 0.5513\\

    \bottomrule
  \end{tabular}}
  \caption{The QA performance and perception of factual knowledge boundaries of LLMs on Natural Questions(NQ) dataset and HotpotQA dataset. Bold denotes the highest score across all the models. Conserv. and Overconf. stand for Conservativeness and Overconfidence respectively.}
  \label{tab:Overall results}
\end{table*}

\paragraph{Metrics.} All the test samples are categorized into four parts based on whether the answer is correct or the model expresses uncertainty and we use $N=N_{cc}+N_{cu}+N_{ic}+N_{iu}$ to represent the total number of test samples, as we can see from Table~\ref{tab:metrics}. We use \textit{accuracy} and \textit{uncertain rate} (Unc-rate for short) to measure the model's QA performance and confidence level respectively. Accuracy be formatted as:
\begin{equation}
    \textit{Accuracy} = \frac{N_{cc}+N_{cu}}{N}    
\end{equation}
which considers a response correct if it contains the ground-truth answer. Uncertain rate is used to represent the proportion of responses where the model express uncertainty and can be formatted as:
\begin{equation}
    \textit{Unc-rate} = \frac{N_{cu}+N_{iu}}{N}
\end{equation}
A lower uncertain rate indicates greater confidence. 
To directly assess the model's perception of knowledge boundaries, we propose \textit{Overconfidence}, \textit{Conservativeness}, and \textit{Alignment} to evaluate the model's extent of overconfidence, conservativeness, and overall perception level, respectively. \textit{Overconfidence} is used to compute the proportion of samples where the model is confident but the response is incorrect. The format is:
\begin{equation}
    \textit{Overconfidence} = \frac{N_{ic}}{N}
\end{equation}

\noindent \textit{Conservativeness} measures the proportion of samples where the model expresses uncertainty but the response is correct and can be formulated as:
\begin{equation}
     \textit{Conservativeness} = \frac{N_{cu}}{N}
\end{equation}

\noindent Alignment is computed by the proportion of samples where the confidence of the model matches the correctness of the response and the format is:
\begin{equation}
    \textit{Alignment} = \frac{N_{cc}+N_{iu}}{N}
\end{equation}


\section{Perception of Factual Knowledge Boundaries in LLMs \label{sec: current state}}
In this section, we select a broad range of representative LLMs and use the vanilla prompt (See Figure~\ref{fig:vanilla template} in Appendix~\S\ref{sec: template}) to test their QA performance and the perceptual level of factual knowledge boundaries. Instead of indirectly characterizing the LLMs' perception of their knowledge boundaries as done by~\citet{ren2023investigating}, we measure this by precise metrics and show the degree of overconfidence and conservativeness. We can observe the overall results in Table \ref{tab:Overall results}. It shows: 
1) The alignment between the QA performance and confidence of LLMs is not high and all the models exhibit overconfidence, even the most powerful model GPT-4. For example, on NQ, GPT-4 can only answer less than 49\% of the questions correctly, yet falsely confirms its answers as wrong in 18.94\% of the cases.
2) Overconfidence is much more severe than the Conservativeness, indicating that the unclear perception of knowledge boundaries is mainly caused by overconfidence.
3) There is no clear correlation between accuracy and the perception of knowledge boundaries. In other words, models with higher accuracy can have lower alignment, e.g., GPT-Instruct versus ChatGPT on both datasets. This suggests that further training on dialogue data may enhance the perception of knowledge boundaries but decrease QA performance.

\begin{figure}[htbp]
  \centering
    \includegraphics[width=0.5\textwidth]{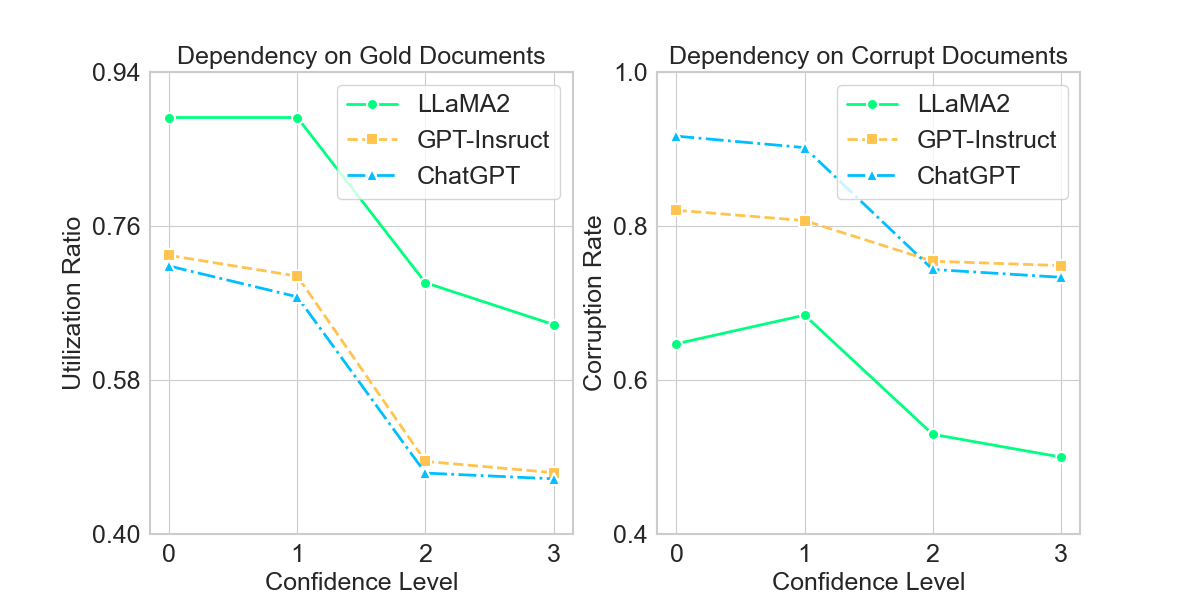}
  \caption{Correlation between certainty and reliance on external information. For LLaMA2, the samples in level 0 and level 1 are nearly identical.}
  \label{fig:dependency}
  \vspace{-0.2cm}
\end{figure}

\section{Correlation between Certainty and Reliance on External Information}
Under retrieval augmentation, we need to know when LLMs show uncertainty to a question, whether they will leverage the provided external information. In this section, we investigate whether LLMs tend to rely on the documents when expressing uncertainty and how the models' confidence levels affect their reliance. 

\subsection{Experimental Setup}
 We guide the model to output its certainty in answering a question correctly using two different prompts and categorize the confidence into four levels based on these two responses. We obtain the certainty $c$ and $\hat{c}$ using the vanilla prompt (See Figure~\ref{fig:vanilla template} in Appendix~\S\ref{sec: template}) and the Punish+Explain method which we propose in Section~\S\ref{sec: alignment enhancement}, respectively. If the model expresses uncertainty twice, it indicates a lack of confidence, whereas two expressions of certainty indicate high confidence. The four confidence levels are delineated as follows: Level 0: $c=0, \hat{c}=0$; Level 1: $c=0$; Level 2: $c=1$; Level 3: $c=1, \hat{c}=1$. Confidence levels increase from level 0 to level 4.
 
We investigate the relationship focusing on two types of supporting documents: \textbf{Gold Documents}: where the ground-truth document provided by DPR~\citep{karpukhin-etal-2020-dense} is used for augmentation. There are 1691 questions with gold documents. \textbf{Corrupt Documents} which are identical to gold documents except that correct answers are replaced with ``Tom''. 

We ask the model to decide whether to rely on its internal knowledge or the document for the answer on its own (See Figure~\ref{fig:ra template}). We test the relationship across three models (i.e., LLaMA2, GPT-Instruct, and ChatGPT) and evaluate the results by two metrics. \textbf{Utilization Ratio}: For a given question $q$ and the document $d$, along with the response $a$ without augmentation, and the response $\hat{a}$ with augmentation. If the $\text{Overlap}(\hat{a}, d) - \text{Overlap}(a, d) > \gamma$ where $\gamma$ is the threshold, we infer that the model relies on the document. In this paper, we set $\gamma=0$. 
\textbf{Corruption Rate}: Percentage of questions where $a$ is right but $\hat{a}$ is wrong.
Utilization ratio is used for gold documents and corruption rate is used for wrong documents. Relying on the gold document does not guarantee a correct answer, as the model may refer to other parts of the document. Therefore, we consider the increase in overlap between the answer and the document as an indicator. However, there is a high probability of generating incorrect answers when relying on the corrupt document. Therefore, if the model generates incorrect answers to questions that it originally could have answered correctly, we consider it to rely on the document.

\subsection{Results and Analysis}
The results are shown in Figure \ref{fig:dependency}. We observe that all the models exhibit a decrement in document dependency as the confidence increases. It indicates LLMs tend to rely more on external documents when they express uncertainty. The overall dependency on the documents is quite high, regardless of whether the documents contain the correct answers. This implies that LLMs tend to trust the input content, making it indispensable to be prudent when leveraging retrieval augmentation, especially when the retriever can have poor performance. This also emphasizes the importance of adaptive retrieval augmentation. \looseness=-1

\begin{table*}[h]
\centering
\scalebox{0.655}{
    \begin{tabular}{llcccccccccc}
    \toprule
     & & \multicolumn{5}{c}{NQ} & \multicolumn{5}{c}{HotpotQA} \\
     \cmidrule(lr){3-7} \cmidrule(lr){8-12}
     \textbf{Model} & \textbf{Strategy} & \textbf{Unc-rate} & \textbf{Acc} & \textbf{Conserv.} & \textbf{Overconf.} & \textbf{Alignment} & \textbf{Unc-rate} & \textbf{Acc} & \textbf{Conserv.} & \textbf{Overconf.} & \textbf{Alignment} \\
     \midrule
     \multirow{7}{*}{ LLaMA2 } & Vanilla & 0.1684 & 0.2986 & 0.0161 & 0.5490 & 0.4349 & 0.4484 & 0.1186 &  0.0230 & 0.4560 & 0.5209 \\
     & Punish & 0.6922 & 0.2277 & 0.1028 & 0.1787 & \textbf{0.7144} & 0.7911 & 0.0907 & 0.0453 & 0.1635 & 0.7912 \\
      & Challenge & \textbf{0.9737} & 0.2986 & \textbf{0.2898} & 0.0174 & 0.6928 & \textbf{0.9825} & 0.1186 & \textbf{0.1160} & 0.0150 & \textbf{0.8690}\\
     & Step-by-Step & 0.3152 & 0.2914 & 0.0371 & 0.4852 & 0.5324 & 0.5009 & 0.1144 & 0.0241 & 0.3998  & 0.5762 \\
     \cmidrule(lr){2-12}
     & Generate & 0.0632 & \textbf{0.3413} & 0.0039 & \textbf{0.5995} & 0.3967 & 0.2718 & \textbf{0.1571} & 0.0096 & \textbf{0.5807} & 0.4096\\
     & Explain & 0.1255 & 0.3332 &  0.0152 & 0.5565 & 0.4283 & 0.4601 & 0.1400 & 0.0237 & 0.4237 & 0.5526 \\
     & Punish+Explain & 0.5080 & 0.2640 &  0.0637 & 0.2917 & 0.6446 & 0.7143 & 0.1161 & 0.0478 & 0.2174 & 0.7348\\
     \midrule
     \multirow{7}{*}{ GPT-Instruct } & Vanilla & 0.1900 & 0.4003 &  0.0346 & 0.4444  & 0.5211 & 0.2188 & 0.2330 & 0.0144 & 0.5626 & 0.4230  \\
     & Punish & 0.2413 & 0.3970 & 0.0454 & 0.4072 & 0.5474  & 0.2522 & 0.2311 & 0.0186 & 0.5353 & 0.4460\\
    & Challenge & \textbf{0.7934} & 0.4003 & \textbf{0.2909} & 0.0972  & \textbf{0.6119} & \textbf{0.8212} & 0.2330 & \textbf{0.1622} & 0.1080 & \textbf{0.7299}\\
     & Step-by-Step & 0.2100 & 0.3798 &  0.0321 & 0.4424  & 0.5255 & 0.1854 & 0.2222 & 0.0132 & 0.6057 & 0.3811\\
     \cmidrule(lr){2-12}
     & Generate & 0.0670 & 0.4349 & 0.0102 & \textbf{0.5083} & 0.4814 & 0.1086 & 0.2503 & 0.0073 & \textbf{0.6484} & 0.3443\\
     & Explain & 0.1560 & \textbf{0.4499} & 0.0255 & 0.4196  & 0.5548 & 0.1651 & \textbf{0.2938} & 0.0136 & 0.5546 & 0.4318\\
     
     & Punish+Explain & 0.2100 & 0.4391 & 0.0371 & 0.3880 & 0.5748 & 0.2083 & 0.2813 & 0.0115 & 0.5219 & 0.4665\\
     \midrule
     \multirow{7}{*}{ ChatGPT } & Vanilla & 0.2917 & 0.3850 &  0.0557 & 0.3789  & 0.5654 & 0.5679 & 0.1951 & 0.0376 & 0.2747 & 0.6877  \\
     & Punish & 0.4086 & 0.3734 &  0.0886 & 0.3066  & 0.6047 & 0.5862 & 0.1854 & 0.0393 & 0.2677 & \textbf{0.6929} \\
      & Challenge & \textbf{0.8875} & 0.3850 & \textbf{0.3714} & 0.0989 & 0.5296 & \textbf{0.8710} & 0.1951 & \textbf{0.1880} & 0.1229 & 0.6882 \\
     & Step-by-Step & 0.3457 & 0.3823 & 0.0779 & 0.3499  & 0.5723 & 0.5479 & 0.1901 & 0.0349 & 0.2969 & 0.6682 \\
     \cmidrule(lr){2-12}
     & Generate & 0.1931 & 0.4224 & 0.0244 & \textbf{0.4089} & 0.5668 & 0.3220 & 0.2267 & 0.0080 & \textbf{0.4592} & 0.5328 \\
     & Explain & 0.2327 & \textbf{0.4424} &  0.0471 & 0.372  & 0.5809 & 0.4203 & \textbf{0.2562} & 0.0303 & 0.3538 & 0.6158\\
     
     & Punish+Explain & 0.2927 & 0.4327 & 0.0573 & 0.3322 & \textbf{0.6102} & 0.4616 & 0.2559 & 0.0344 & 0.3169 & 0.6487 \\
     \midrule
     \multirow{4}{*}{ GPT-4* } & Vanilla & 0.1360 & 0.5920 &  0.0280 & \textbf{0.3000}  & 0.6720 & 0.2680 & 0.4040 & 0.0400 & \textbf{0.4220} & 0.5560 \\
     & Punish & 0.2100 & 0.5376 &  0.0660 & 0.2800  & 0.6540 & 0.3500 & 0.4140 & 0.0760 & 0.3260 & 0.5920 \\
     \cmidrule(lr){2-12}
     & Explain & 0.2080 & \textbf{0.6660} &  0.0800 & 0.2060  & \textbf{0.7140} & 0.3820 & \textbf{0.5160} & 0.0980 & 0.2540  & 0.6580 \\
     & Punish+Explain & \textbf{0.3320} & 0.6500 & \textbf{0.1560} & 0.1740 & 0.6700 & \textbf{0.5180} & 0.4840 & \textbf{0.1460} & 0.1840 & \textbf{0.6660} \\
    \bottomrule
\end{tabular}}
\caption{Performance of different methods on Natural Question(NQ) and HotpotQA datasets. Bold denotes the highest scores across all the methods for each model. The model marked with * indicates that the results are based on the sampled data. Due to budget limit, we only employ the most efficient methods for experiments on GPT-4.}
\label{tab:evaluation of strategies on NQ}
\end{table*}

\begin{figure*}[htbp]
  \centering

  \subfigure[Overconfidence on NQ]{
    \includegraphics[width=0.45\textwidth]{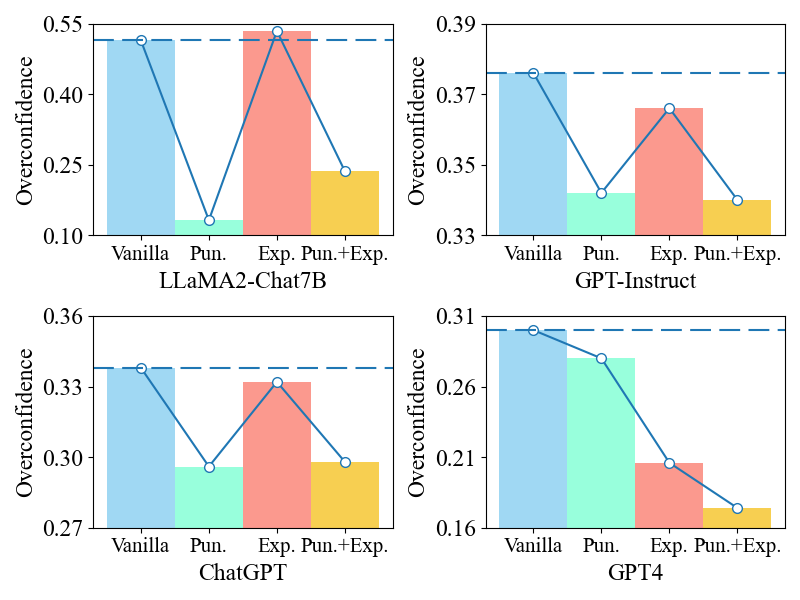}
    \label{fig:overconf-nq}
    \centering
  }
\hspace{.12in}
  \subfigure[Overconfidence on HotpotQA]{
    \includegraphics[width=0.45\textwidth]{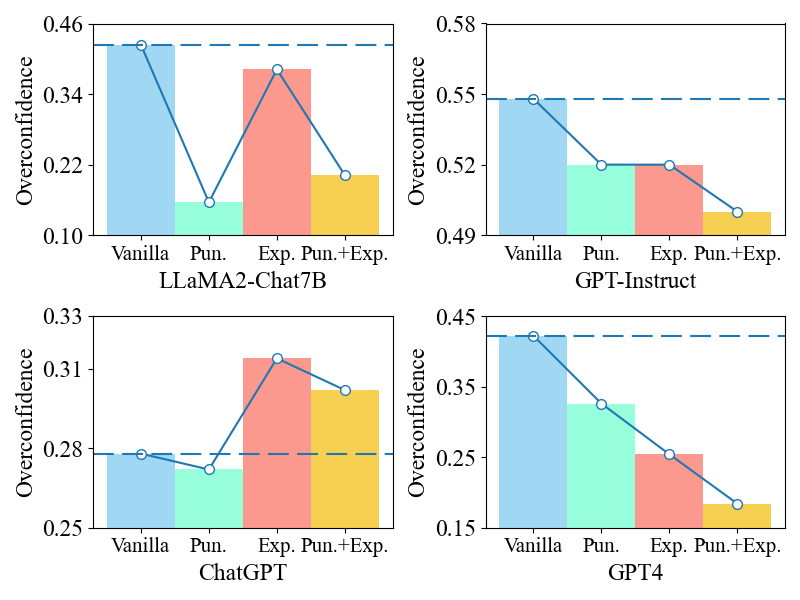}
    \label{fig:overconf-hq}
    \centering
  }

  \caption{The overconfidence for each model under different strategies on the 500 sampled data. Pun. represents the Punish method, Exp. represents the Explain method, and Pun.+Exp. represents the Punish+Explain method.}
  \label{fig:overconf}
\end{figure*}
\section{Alignment Enhancement \label{sec: alignment enhancement}}
As discovered in Section~\S\ref{sec: current state}, the poor perception of knowledge boundaries in LLMs is mainly caused by their overconfidence. Therefore, we enhance the perception of knowledge boundaries by mitigating overconfidence. This can be done from two perspectives: urging LLMs to be prudent and improving their ability to provide correct answers.

\subsection{Mitigating Overconfidence}

We designed prompts from two perspectives with the aim of mitigating overconfidence.

\paragraph{Methods aimed at urging LLMs to be prudent.} We design three types of prompts to reduce the confidence. \textbf{1. Punish}: We add ``\textit{You will be punished if the answer is not right but you say certain}" to the prompt, encouraging the model to be prudent. \textbf{2. Challenge}: We challenge the correctness of the generated answer and force the model to express more uncertainty. \textbf{3. Think Step by Step}: The ``Think step by step'' approach has been proven to be an effective way to enhance the reasoning ability~\citep{kojima2022large}. Therefore, we explicitly ask the model to think step by step, answering the question first and outputting the confidence in the next step. We hope the model can recognize its overconfidence when asked to think step by step. 

\paragraph{Methods aimed at enhancing QA performance.} We design two methods to enhance the accuracy. \textbf{1. Generate}: LLMs can generate high-quality documents on their own, thereby assisting in generating accurate answers~\citep{yu2022generate}. We ask the model to generate a short document that aids in answering the question, ultimately bolstering the accuracy of the response. \textbf{2. Explain}: In addition to generating auxiliary information before providing the answer, we may also obtain more reliable results by asking the model to explain the reason about its answer. This may mitigate the risk of the producing incorrect responses lacking reasonable explanation.

 To combine the concepts of being prudent and enhancing QA performance, we merge the Punish and Explain methods into a single approach, called \textbf{Punish+Explain}. We can find all the proposed prompts in Appendix~\S\ref{sec: template}.

\subsection{Results and Analysis \label{sec: results for enhancing alignment}}
 We can find the performance of different strategies on NQ and HotpotQA in Table~\ref{tab:evaluation of strategies on NQ}. We do not investigate Vicuna because it is too confident compared to the other models. More details can be found in Appendix~\S\ref{sec:vicuna}. Here are our observations: 
1) All the methods aimed at urging models to be prudent result in an increased proportion of uncertain responses as expected. The Challenge method dramatically increases the proportion of uncertain responses, achieving the lowest level of overconfidence and the highest degree of conservativeness among all the methods. This suggests that LLMs tend to trust the input and undermine their own judgments, leading to excessive conservativeness. In contrast, the Punish method weakens overconfidence without making the model overly conservative, which typically leads to an improvement in alignment. The Think Step by Step method reduces the degree of overconfidence on NQ dataset but exacerbates it on HotpotQA dataset. Thus, this method is not particularly effective. Moreover, the Punish and Step by Step methods may result in a slight performance decrease.

2) All the methods aimed at enhancing QA performance lead to higher answer accuracy. Generate method produces the highest overconfidence scores among all the methods. The possible reason may be that LLMs generate documents that aid in answering questions as expected. However, relying on self-generated documents leads LLMs to believe their answers are correct. The difference lies in the fact that the Explain method typically diminishes overconfidence and maintains comparable or even lower conservativeness levels, thereby enhancing the perception of knowledge boundaries for LLMs. The overconfidence of ChatGPT is the lowest on the HQ dataset, making it difficult to further reduce through methods aimed at enhancing accuracy.  \looseness=-1

3) The Punish method is highly effective for LLaMA2, while the Explain method is highly effective for GPT-4. This may be because LLaMA2 exhibits severe overconfidence and has weaker generation capabilities, making the Punish method more effective. On the other hand, GPT-4 shows lower level of overconfidence. Given its strong generation capabilities, the Explain method significantly improves accuracy and reduces overconfidence. To combine the concepts of urging models to be prudent and enhancing QA performance, we merge the Punish and the Explain methods into a single approach, called Punish+Explain. Compared to the individual methods, this approach consistently enhances alignment without compromising accuracy.

We conduct experiments on the other models using the same 500 sampled data utilized for GPT-4. To facilitate clarity, we illustrate the effects of various strategies on overconfidence in Figure~\ref{fig:overconf}, while comprehensive details are provided in Table~\ref{tab:evaluation of strategies on sampled data} in Appendix~\S\ref{sec:results on the sampled data}. The conclusions on the 500 samples align with those from the full dataset.

\section{Adaptive Retrieval Augmentation \label{sec:adaptive retrieval augmentation}}
Our work focuses on determining when to conduct retrieval rather than triggering retrieval all the time and enhancing the ability of LLMs to leverage a document of unknown quality. In this section, we introduce the methods proposed in Section~\S\ref{sec: alignment enhancement} to adaptive retrieval augmentation.

\subsection{Experimental Settings}
We conduct retrieval augmentation under two settings. \textbf{Static retrieval augmentation}: We enable retrieval augmentation for all the questions. \textbf{Adaptive retrieval augmentation}: We adaptively enable retrieval augmentation when the model believes that it cannot answer the question based on its internal knowledge based on the four prompts: Vanilla, Punish, Explain, and Punish+Explain. 

We do not conduct adaptive retrieval augmentation on Vicuna because Vicuna is notably more confident compared to the other models, resulting in a very low proportion of uncertainty. Therefore, applying adaptive retrieval augmentation to Vicuna hardly triggers any enhancement. More details can be found in Appendix~\S\ref{sec:vicuna}.

\begin{table}[htbp]
    \centering
    \begin{tabular}{cccc}
    \toprule
         Dataset & Sparse & Dense & Gold\\
         \midrule
          NQ  & 0.26 & 0.61 & 1.00\\
         HotpotQA & 0.33 & 0.32 & 1.00\\ 
         \bottomrule
    \end{tabular}
    \caption{Precision@1 results for different retrievers}
    \label{tab:P@1}
\end{table}
\paragraph{Retrievers.} We consider three types of supporting documents including Sparse documents retrieved through Sparse retrieval~\citep{robertson2009probabilistic}, Dense documents retrieved through Dense retrieval~\citep{guo2022semantic, ni2023comparative} and Gold documents which contain the correct answer. Dense documents represent the practical usage and the other two respectively represent the lower and upper bounds of the actual situation. The knowledge source is a Wikipedia dump provided by DPR~\citep{karpukhin-etal-2020-dense}. Following the previous study~\citep{ren2023investigating}, we use RocketQAv2~\cite{ren-etal-2021-rocketqav2} as the dense retriever to find semantically relevant documents for each question. For the sparse retriever, we use BM25\citep{yang2017anserini} to retrieve relevant documents from the lexical level. We obtain gold documents which contain the correct for NQ like \citet{karpukhin-etal-2020-dense} and for HQ, we get the gold documents as \citet{ren2023investigating} did. To focus on the effect of model perception of knowledge boundaries on adaptive retrieval augmentation, for simplicity, we only provide LLMs the top-1 document and the retrieval performance can be seen in Table~\ref{tab:P@1}. As described in Section~\S\ref{sec: alignment enhancement}, the conclusions on the sampled data remain consistent with those on the full dataset. Therefore, for the budget concern, we only conduct experiments on the 500 sampled data. The prompt used for retrieval augmentation can be found in Figure~\ref{fig:ra template}.

\begin{table*}[h]
\scalebox{0.75}{
\begin{tabular}{cccccccccccc}
\toprule
 & & \multicolumn{5}{c}{NQ} & \multicolumn{5}{c}{HotpotQA} \\
     \cmidrule(lr){3-7} \cmidrule(lr){8-12}
  \textbf{Model} & \textbf{Retrieval}  & \textbf{Static} & \textbf{Vanilla} & \textbf{Punish} & \textbf{Explain} & \textbf{Pun.+Exp.} & \textbf{Static} & \textbf{Vanilla} & \textbf{Punish} & \textbf{Explain} & \textbf{Pun.+Exp.}\\
\midrule
  \multirow{5}{*}{ LLaMA2} & RA Rate & 100\% & 14.6\% & 71.4\% & 9.2\% & 51.8\% & 100\% & 44.8\% & 78.4\% & 46.2\% & 70.2\% \\
  \cmidrule(lr){2-12}
  & None & 0.352 & 0.352 & 0.276 & \textbf{0.382} & 0.316 & 0.160 & 0.160 & 0.138 & \textbf{0.186} & 0.172  \\ 
  & Sparse & 0.256 & 0.370 & 0.316 & \textbf{0.390} & 0.356  & \textbf{0.334} & 0.270 & 0.310 & 0.298 & 0.314 \\
  & Dense & \textbf{0.534} & 0.414 & 0.522 & 0.418 & 0.494 & 0.288 & 0.244 & 0.276 & 0.276 & \textbf{0.292}  \\
  & Gold & \textbf{0.774} & 0.460 & 0.706 & 0.448 & 0.642  & \textbf{0.516} & 0.370 & 0.468 & 0.412 & 0.474 \\
 \midrule
   \multirow{5}{*}{ GPT-Instruct } & RA Rate &100\% & 16.6\% & 21.4\% & 13.4\% & 16.8\% & 100\% & 18.0\% & 20.6\% & 12.0\% & 16.2\% \\ 
   \cmidrule(lr){2-12}
   & None & 0.496 & 0.496 & 0.486 & 0.522 & \textbf{0.528} & 0.294 & 0.294 & 0.302 & \textbf{0.378} & 0.354 \\ 
  & Sparse & 0.282 & 0.474 & 0.476 & \textbf{0.516} & 0.512 & 0.344 & 0.312 & 0.316 & \textbf{0.390} & 0.374 \\
  & Dense & 0.538 & 0.518 & 0.520 & 0.538 & \textbf{0.554} & 0.324 & 0.306 & 0.306 & \textbf{0.378} & 0.362 \\
  & Gold & \textbf{0.816} & 0.588 & 0.614 & 0.602 & 0.620 & \textbf{0.568} & 0.354 & 0.364 & 0.422 & 0.418 \\
 \midrule
  \multirow{5}{*}{ ChatGPT } & RA Rate & 100\% & 25.4\% & 33.2\% & 17.8\% & 22.6\% & 100\% & 52.6\% & 52.6\% & 39.4\% & 40.6\% \\ 
  \cmidrule(lr){2-12}
  & None & 0.468 & 0.468 & 0.456 & 0.530 & \textbf{0.536} & 0.240 & 0.240 & 0.236 & \textbf{0.326} & 0.326 \\ 
  & Sparse & 0.228 & 0.448 & 0.422 & \textbf{0.510} & 0.504 & 0.276 & 0.300 & 0.300 & \textbf{0.360} & 0.346 \\
  & Dense & 0.506 & 0.490 & 0.488 & 0.550 & \textbf{0.556} & 0.238 & 0.276 & 0.266 & \textbf{0.344} & 0.336  \\
  & Gold & \textbf{0.800} & 0.602 & 0.630 & 0.616 & 0.646 & 0.406 & 0.352 & 0.350 & 0.404 & \textbf{0.412} \\
 \midrule
  \multirow{5}{*}{ GPT-4 } & RA Rate & 100\% & 13.6\% & 21.0\% & 20.8\% & 33.2\% & 100\% & 26.8\% & 35.0\% & 38.2\% & 51.8\% \\ 
  \cmidrule(lr){2-12}
  & None & 0.592 & 0.592 & 0.538 & \textbf{0.666} & 0.650 & 0.404 & 0.404 & 0.414 & \textbf{0.516} & 0.484 \\ 
  & Sparse & 0.572 & 0.610 & 0.600 & \textbf{0.664} & 0.634 & 0.546 & 0.464 & 0.478 & 0.566 & \textbf{0.568} \\
  & Dense & \textbf{0.698} & 0.622 & 0.624 & 0.688 & 0.676 & 0.510 & 0.458 & 0.464 & \textbf{0.540} & 0.528 \\
  & Gold & \textbf{0.866} & 0.676 & 0.680 & 0.756 & 0.764 & \textbf{0.644} & 0.500 & 0.530 & 0.616 & 0.620 \\
 \bottomrule
\end{tabular}}
\caption{Accuracy of each model under different strategies for retrieval augmentation. RA Rate represents the proportion of triggering retrieval augmentation. None represents accuracy without retrieval augmentation. Bold indicates the best performance under the current retrieval setting. The results are all on the sampled data.} 
\label{tab:Results of adaptive retrieval augmentation}
\end{table*}

\subsection{Results and Analysis}
Table \ref{tab:Results of adaptive retrieval augmentation} illustrates the accuracy of answers under each strategy on NQ and HotpotQA in Table~\ref{tab:evaluation of strategies on NQ}. Our findings are as follows:

1) When using a gold document for augmentation, static augmentation achieves the highest accuracy in almost all the cases. It shows documents containing the answers often help answer the questions. For adaptive retrieval augmentation, in most cases, the Punish+Explain method achieves the best results because it consistently enhances alignment without compromising QA performance. The best performance obtained in adaptive retrieval augmentation does not differ significantly from the static augmentation, and in some cases, it even achieves comparable or better performance, while utilizing only a minimal number of retrieval attempts. For example, ChatGPT achieves the best performance on HotpotQA by utilizing 40.6\% of retrievals under Punish+Explain strategy.

2) Our adaptive retrieval augmentation makes LLMs more robust to documents that may not help. When utilizing documents retrieved through the sparse retriever, it is observed that static augmentation often leads to performance degradation on NQ. This is because LLMs perform well on these questions, and providing low-quality documents can mislead the models. In contrast, adaptive retrieval augmentation can reduce performance loss or even lead to improvement. The highest accuracy is often achieved under the Explain strategy because this method inherently enhances performance and has a relatively small uncertainty rate. 

3) In real search scenarios, Explain and Punish+Explain strategies are more efficient than the static augmentation when documents contribute to improving accuracy. We observe that employing sparse retrieval documents for static augmentation on NQ and utilizing both sparse and dense retrieval documents for static augmentation on HQ frequently result in performance enhancements. This suggests that these documents typically provide assistance. Compared to static augmentation, adaptive retrieval augmentation consistently achieves comparable or even superior performance under the Explain and Punish+Explain strategies, while requiring fewer retrieval augmentation attempts. \looseness=-1

Compared to static RA, although adaptive RA requires two rounds of inference when augmentation is needed, the cost of the initial round of uncertainty perception is much lower due to the significantly shorter input lengths compared to the retrieved documents or passages. Therefore, our strategies generally save on overhead.

\section{Conclusion}
In this paper, we explore the direction of effective and efficient adaptive retrieval augmentation by enhancing LLMs' perception of their knowledge boundary.
First, we propose several metrics to quantitatively measure LLMs' perception of knowledge boundaries and find that overconfidence is the primary reason for the unsatisfactory perception of knowledge boundaries.
To see whether LLMs' certainty has an impact on their reliance on external knowledge, we investigate and find that the more LLMs are uncertain about their internal knowledge, the more they rely on external knowledge. 
We further probe to enhance the LLMs' perception from two perspectives and find that equipped with these methods, LLMs can achieve comparable or even better performance of retrieval augmentation with much fewer retrieval calls.

\section*{Limitations}
First, we divide model's confidence about its answer into two components, without delving into finer granularity.  
Second, our methods mitigate LLMs' overconfidence through prompts, making it difficult to significantly adjust models with excessive overconfidence (i.e., Vicuna-v1.5-7B). For open-source models, there may be better training methods available.  
Additionally, we only focus on LLMs' perception levels of their factual knowledge boundaries. LLMs' perception of knowledge boundaries regarding different types of knowledge remain to be studied.

\section*{Ethics Statement}
We approach ethics with great care. In this paper, all the datasets we use are open-source, and the models we employ are either open-source or widely used. Furthermore, the methods we propose do not induce the model to output any harmful information.

\section*{Acknowledgements}
This work was funded by the National Natural Science Foundation of China (NSFC) under Grants No. 62302486, the Innovation Project of ICT CAS under Grants No. E361140, the CAS Special Research Assistant Funding Project, the Lenovo-CAS Joint Lab Youth Scientist Project, the project under Grants No. JCKY2022130C039, and the Strategic Priority Research Program of the CAS under Grants No. XDB0680102.

\bibliography{anthology,custom}

\begin{thebibliography}{34}
\expandafter\ifx\csname natexlab\endcsname\relax\def\natexlab#1{#1}\fi

\bibitem[{Bi et~al.(2024{\natexlab{a}})Bi, Liu, Mei, Wang, Ji, and Cheng}]{bi2024decoding}
Baolong Bi, Shenghua Liu, Lingrui Mei, Yiwei Wang, Pengliang Ji, and Xueqi Cheng. 2024{\natexlab{a}}.
\newblock Decoding by contrasting knowledge: Enhancing llms' confidence on edited facts.
\newblock \emph{arXiv preprint arXiv:2405.11613}.

\bibitem[{Bi et~al.(2024{\natexlab{b}})Bi, Liu, Wang, Mei, and Cheng}]{bi2024lpnl}
Baolong Bi, Shenghua Liu, Yiwei Wang, Lingrui Mei, and Xueqi Cheng. 2024{\natexlab{b}}.
\newblock Lpnl: Scalable link prediction with large language models.
\newblock \emph{arXiv preprint arXiv:2401.13227}.

\bibitem[{Brown et~al.(2020)Brown, Mann, Ryder, Subbiah, Kaplan, Dhariwal, Neelakantan, Shyam, Sastry, Askell et~al.}]{brown2020language}
Tom Brown, Benjamin Mann, Nick Ryder, Melanie Subbiah, Jared~D Kaplan, Prafulla Dhariwal, Arvind Neelakantan, Pranav Shyam, Girish Sastry, Amanda Askell, et~al. 2020.
\newblock Language models are few-shot learners.
\newblock \emph{Advances in neural information processing systems}, 33:1877--1901.

\bibitem[{Cheng et~al.(2021)Cheng, Shen, Liu, He, Chen, and Gao}]{cheng2021unitedqa}
Hao Cheng, Yelong Shen, Xiaodong Liu, Pengcheng He, Weizhu Chen, and Jianfeng Gao. 2021.
\newblock Unitedqa: A hybrid approach for open domain question answering.
\newblock \emph{arXiv preprint arXiv:2101.00178}.

\bibitem[{Fan et~al.(2024)Fan, Li, Zou, Li, He, Chern, Hu, and Liu}]{fan2024reformatted}
Run-Ze Fan, Xuefeng Li, Haoyang Zou, Junlong Li, Shwai He, Ethan Chern, Jiewen Hu, and Pengfei Liu. 2024.
\newblock Reformatted alignment.
\newblock \emph{arXiv preprint arXiv:2402.12219}.

\bibitem[{Guo et~al.(2017)Guo, Pleiss, Sun, and Weinberger}]{guo2017calibration}
Chuan Guo, Geoff Pleiss, Yu~Sun, and Kilian~Q Weinberger. 2017.
\newblock On calibration of modern neural networks.
\newblock In \emph{International conference on machine learning}, pages 1321--1330. PMLR.

\bibitem[{Guo et~al.(2022)Guo, Cai, Fan, Sun, Zhang, and Cheng}]{guo2022semantic}
Jiafeng Guo, Yinqiong Cai, Yixing Fan, Fei Sun, Ruqing Zhang, and Xueqi Cheng. 2022.
\newblock Semantic models for the first-stage retrieval: A comprehensive review.
\newblock \emph{ACM Transactions on Information Systems (TOIS)}, 40(4):1--42.

\bibitem[{Guu et~al.(2020)Guu, Lee, Tung, Pasupat, and Chang}]{guu2020retrieval}
Kelvin Guu, Kenton Lee, Zora Tung, Panupong Pasupat, and Mingwei Chang. 2020.
\newblock Retrieval augmented language model pre-training.
\newblock In \emph{International conference on machine learning}, pages 3929--3938. PMLR.

\bibitem[{Izacard and Grave(2020)}]{izacard2020leveraging}
Gautier Izacard and Edouard Grave. 2020.
\newblock Leveraging passage retrieval with generative models for open domain question answering.
\newblock \emph{arXiv preprint arXiv:2007.01282}.

\bibitem[{Jiang et~al.(2021)Jiang, Araki, Ding, and Neubig}]{jiang2021can}
Zhengbao Jiang, Jun Araki, Haibo Ding, and Graham Neubig. 2021.
\newblock How can we know when language models know? on the calibration of language models for question answering.
\newblock \emph{Transactions of the Association for Computational Linguistics}, 9:962--977.

\bibitem[{Karpukhin et~al.(2020)Karpukhin, Oguz, Min, Lewis, Wu, Edunov, Chen, and Yih}]{karpukhin-etal-2020-dense}
Vladimir Karpukhin, Barlas Oguz, Sewon Min, Patrick Lewis, Ledell Wu, Sergey Edunov, Danqi Chen, and Wen-tau Yih. 2020.
\newblock \href {https://doi.org/10.18653/v1/2020.emnlp-main.550} {Dense passage retrieval for open-domain question answering}.
\newblock In \emph{Proceedings of the 2020 Conference on Empirical Methods in Natural Language Processing (EMNLP)}, pages 6769--6781, Online. Association for Computational Linguistics.

\bibitem[{Kojima et~al.(2022)Kojima, Gu, Reid, Matsuo, and Iwasawa}]{kojima2022large}
Takeshi Kojima, Shixiang~Shane Gu, Machel Reid, Yutaka Matsuo, and Yusuke Iwasawa. 2022.
\newblock Large language models are zero-shot reasoners.
\newblock \emph{Advances in neural information processing systems}, 35:22199--22213.

\bibitem[{Kwiatkowski et~al.(2019)Kwiatkowski, Palomaki, Redfield, Collins, Parikh, Alberti, Epstein, Polosukhin, Devlin, Lee et~al.}]{kwiatkowski2019natural}
Tom Kwiatkowski, Jennimaria Palomaki, Olivia Redfield, Michael Collins, Ankur Parikh, Chris Alberti, Danielle Epstein, Illia Polosukhin, Jacob Devlin, Kenton Lee, et~al. 2019.
\newblock Natural questions: a benchmark for question answering research.
\newblock \emph{Transactions of the Association for Computational Linguistics}, 7:453--466.

\bibitem[{Lewis et~al.(2020)Lewis, Perez, Piktus, Petroni, Karpukhin, Goyal, K{\"u}ttler, Lewis, Yih, Rockt{\"a}schel et~al.}]{lewis2020retrieval}
Patrick Lewis, Ethan Perez, Aleksandra Piktus, Fabio Petroni, Vladimir Karpukhin, Naman Goyal, Heinrich K{\"u}ttler, Mike Lewis, Wen-tau Yih, Tim Rockt{\"a}schel, et~al. 2020.
\newblock Retrieval-augmented generation for knowledge-intensive nlp tasks.
\newblock \emph{Advances in Neural Information Processing Systems}, 33:9459--9474.

\bibitem[{Liu et~al.(2023)Liu, Zhang, Guo, de~Rijke, Chen, Fan, and Cheng}]{liu2023black}
Yu-An Liu, Ruqing Zhang, Jiafeng Guo, Maarten de~Rijke, Wei Chen, Yixing Fan, and Xueqi Cheng. 2023.
\newblock Black-box adversarial attacks against dense retrieval models: A multi-view contrastive learning method.
\newblock In \emph{Proceedings of the 32nd ACM International Conference on Information and Knowledge Management}, pages 1647--1656.

\bibitem[{Liu et~al.(2024)Liu, Zhang, Guo, de~Rijke, Fan, and Cheng}]{liu2024multi}
Yu-An Liu, Ruqing Zhang, Jiafeng Guo, Maarten de~Rijke, Yixing Fan, and Xueqi Cheng. 2024.
\newblock Multi-granular adversarial attacks against black-box neural ranking models.
\newblock \emph{arXiv preprint arXiv:2404.01574}.

\bibitem[{Mallen et~al.(2023)Mallen, Asai, Zhong, Das, Khashabi, and Hajishirzi}]{mallen-etal-2023-trust}
Alex Mallen, Akari Asai, Victor Zhong, Rajarshi Das, Daniel Khashabi, and Hannaneh Hajishirzi. 2023.
\newblock \href {https://doi.org/10.18653/v1/2023.acl-long.546} {When not to trust language models: Investigating effectiveness of parametric and non-parametric memories}.
\newblock In \emph{Proceedings of the 61st Annual Meeting of the Association for Computational Linguistics (Volume 1: Long Papers)}, pages 9802--9822, Toronto, Canada. Association for Computational Linguistics.

\bibitem[{Minderer et~al.(2021)Minderer, Djolonga, Romijnders, Hubis, Zhai, Houlsby, Tran, and Lucic}]{minderer2021revisiting}
Matthias Minderer, Josip Djolonga, Rob Romijnders, Frances Hubis, Xiaohua Zhai, Neil Houlsby, Dustin Tran, and Mario Lucic. 2021.
\newblock Revisiting the calibration of modern neural networks.
\newblock \emph{Advances in Neural Information Processing Systems}, 34:15682--15694.

\bibitem[{Ni et~al.(2023)Ni, Bi, Guo, and Cheng}]{ni2023comparative}
Shiyu Ni, Keping Bi, Jiafeng Guo, and Xueqi Cheng. 2023.
\newblock A comparative study of training objectives for clarification facet generation.
\newblock In \emph{Proceedings of the Annual International ACM SIGIR Conference on Research and Development in Information Retrieval in the Asia Pacific Region}, pages 1--10.

\bibitem[{Ouyang et~al.(2022)Ouyang, Wu, Jiang, Almeida, Wainwright, Mishkin, Zhang, Agarwal, Slama, Ray et~al.}]{ouyang2022training}
Long Ouyang, Jeffrey Wu, Xu~Jiang, Diogo Almeida, Carroll Wainwright, Pamela Mishkin, Chong Zhang, Sandhini Agarwal, Katarina Slama, Alex Ray, et~al. 2022.
\newblock Training language models to follow instructions with human feedback.
\newblock \emph{Advances in Neural Information Processing Systems}, 35:27730--27744.

\bibitem[{Qu et~al.(2020)Qu, Ding, Liu, Liu, Ren, Zhao, Dong, Wu, and Wang}]{qu2020rocketqa}
Yingqi Qu, Yuchen Ding, Jing Liu, Kai Liu, Ruiyang Ren, Wayne~Xin Zhao, Daxiang Dong, Hua Wu, and Haifeng Wang. 2020.
\newblock Rocketqa: An optimized training approach to dense passage retrieval for open-domain question answering.
\newblock \emph{arXiv preprint arXiv:2010.08191}.

\bibitem[{Ren et~al.(2021)Ren, Qu, Liu, Zhao, She, Wu, Wang, and Wen}]{ren-etal-2021-rocketqav2}
Ruiyang Ren, Yingqi Qu, Jing Liu, Wayne~Xin Zhao, QiaoQiao She, Hua Wu, Haifeng Wang, and Ji-Rong Wen. 2021.
\newblock \href {https://doi.org/10.18653/v1/2021.emnlp-main.224} {{R}ocket{QA}v2: A joint training method for dense passage retrieval and passage re-ranking}.
\newblock In \emph{Proceedings of the 2021 Conference on Empirical Methods in Natural Language Processing}, pages 2825--2835, Online and Punta Cana, Dominican Republic. Association for Computational Linguistics.

\bibitem[{Ren et~al.(2023)Ren, Wang, Qu, Zhao, Liu, Tian, Wu, Wen, and Wang}]{ren2023investigating}
Ruiyang Ren, Yuhao Wang, Yingqi Qu, Wayne~Xin Zhao, Jing Liu, Hao Tian, Hua Wu, Ji-Rong Wen, and Haifeng Wang. 2023.
\newblock Investigating the factual knowledge boundary of large language models with retrieval augmentation.
\newblock \emph{arXiv preprint arXiv:2307.11019}.

\bibitem[{Robertson et~al.(2009)Robertson, Zaragoza et~al.}]{robertson2009probabilistic}
Stephen Robertson, Hugo Zaragoza, et~al. 2009.
\newblock The probabilistic relevance framework: Bm25 and beyond.
\newblock \emph{Foundations and Trends{\textregistered} in Information Retrieval}, 3(4):333--389.

\bibitem[{Shi et~al.(2023)Shi, Min, Yasunaga, Seo, James, Lewis, Zettlemoyer, and Yih}]{shi2023replug}
Weijia Shi, Sewon Min, Michihiro Yasunaga, Minjoon Seo, Rich James, Mike Lewis, Luke Zettlemoyer, and Wen-tau Yih. 2023.
\newblock Replug: Retrieval-augmented black-box language models.
\newblock \emph{arXiv preprint arXiv:2301.12652}.

\bibitem[{Shi et~al.(2024{\natexlab{a}})Shi, Gao, Chen, Feng, Yan, Shi, Yin, Chen, Verberne, and Ren}]{shi2024chain}
Zhengliang Shi, Shen Gao, Xiuyi Chen, Yue Feng, Lingyong Yan, Haibo Shi, Dawei Yin, Zhumin Chen, Suzan Verberne, and Zhaochun Ren. 2024{\natexlab{a}}.
\newblock Chain of tools: Large language model is an automatic multi-tool learner.
\newblock \emph{arXiv preprint arXiv:2405.16533}.

\bibitem[{Shi et~al.(2024{\natexlab{b}})Shi, Gao, Chen, Yan, Shi, Yin, Chen, Ren, Verberne, and Ren}]{shi2024learning}
Zhengliang Shi, Shen Gao, Xiuyi Chen, Lingyong Yan, Haibo Shi, Dawei Yin, Zhumin Chen, Pengjie Ren, Suzan Verberne, and Zhaochun Ren. 2024{\natexlab{b}}.
\newblock Learning to use tools via cooperative and interactive agents.
\newblock \emph{arXiv preprint arXiv:2403.03031}.

\bibitem[{Singh et~al.(2021)Singh, Reddy, Hamilton, Dyer, and Yogatama}]{singh2021end}
Devendra Singh, Siva Reddy, Will Hamilton, Chris Dyer, and Dani Yogatama. 2021.
\newblock End-to-end training of multi-document reader and retriever for open-domain question answering.
\newblock \emph{Advances in Neural Information Processing Systems}, 34:25968--25981.

\bibitem[{Yang et~al.(2017)Yang, Fang, and Lin}]{yang2017anserini}
Peilin Yang, Hui Fang, and Jimmy Lin. 2017.
\newblock Anserini: Enabling the use of lucene for information retrieval research.
\newblock In \emph{Proceedings of the 40th international ACM SIGIR conference on research and development in information retrieval}, pages 1253--1256.

\bibitem[{Yang et~al.(2023)Yang, Chern, Qiu, Neubig, and Liu}]{yang2023alignment}
Yuqing Yang, Ethan Chern, Xipeng Qiu, Graham Neubig, and Pengfei Liu. 2023.
\newblock Alignment for honesty.
\newblock \emph{arXiv preprint arXiv:2312.07000}.

\bibitem[{Yang et~al.(2018)Yang, Qi, Zhang, Bengio, Cohen, Salakhutdinov, and Manning}]{yang2018hotpotqa}
Zhilin Yang, Peng Qi, Saizheng Zhang, Yoshua Bengio, William~W Cohen, Ruslan Salakhutdinov, and Christopher~D Manning. 2018.
\newblock Hotpotqa: A dataset for diverse, explainable multi-hop question answering.
\newblock \emph{arXiv preprint arXiv:1809.09600}.

\bibitem[{Yin et~al.(2023)Yin, Sun, Guo, Wu, Qiu, and Huang}]{yin2023large}
Zhangyue Yin, Qiushi Sun, Qipeng Guo, Jiawen Wu, Xipeng Qiu, and Xuanjing Huang. 2023.
\newblock \href {https://doi.org/10.18653/v1/2023.findings-acl.551} {Do large language models know what they don{'}t know?}
\newblock In \emph{Findings of the Association for Computational Linguistics: ACL 2023}, pages 8653--8665, Toronto, Canada. Association for Computational Linguistics.

\bibitem[{Yu et~al.(2022)Yu, Iter, Wang, Xu, Ju, Sanyal, Zhu, Zeng, and Jiang}]{yu2022generate}
Wenhao Yu, Dan Iter, Shuohang Wang, Yichong Xu, Mingxuan Ju, Soumya Sanyal, Chenguang Zhu, Michael Zeng, and Meng Jiang. 2022.
\newblock Generate rather than retrieve: Large language models are strong context generators.
\newblock \emph{arXiv preprint arXiv:2209.10063}.

\bibitem[{Zhang et~al.(2024)Zhang, Zhang, Guo, de~Rijke, Fan, and Cheng}]{zhang2024large}
Hengran Zhang, Ruqing Zhang, Jiafeng Guo, Maarten de~Rijke, Yixing Fan, and Xueqi Cheng. 2024.
\newblock Are large language models good at utility judgments?
\newblock \emph{arXiv preprint arXiv:2403.19216}.

\end{thebibliography}
\bibliographystyle{acl_natbib}
\appendix
\section{Appendix}
\subsection{Prompts \label{sec: template}}
\begin{figure*}[htbp]
  \centering
    \includegraphics[width=\textwidth]{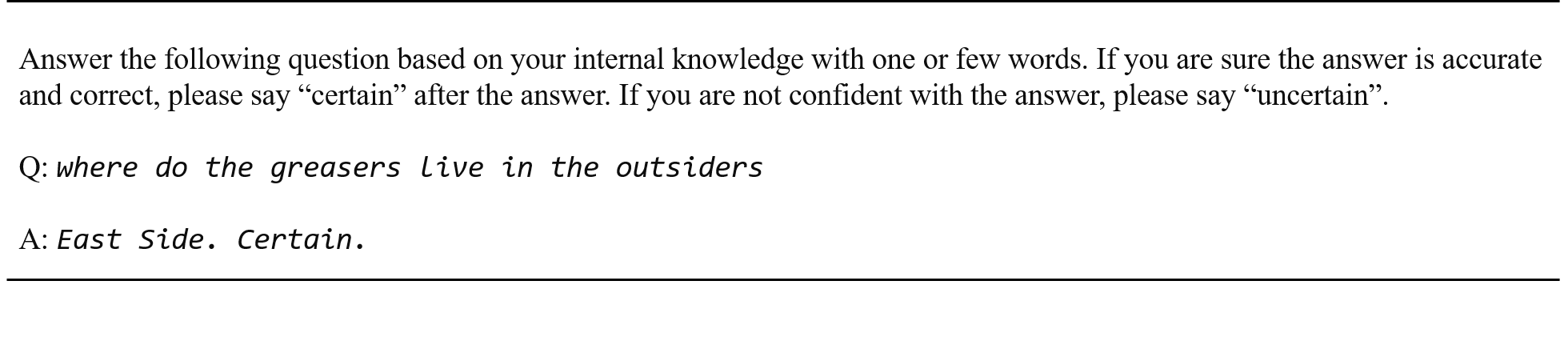}
  \caption{Vanilla template form}
  \label{fig:vanilla template}
\end{figure*}

\begin{figure*}[htbp]
  \centering
    \includegraphics[width=\textwidth]{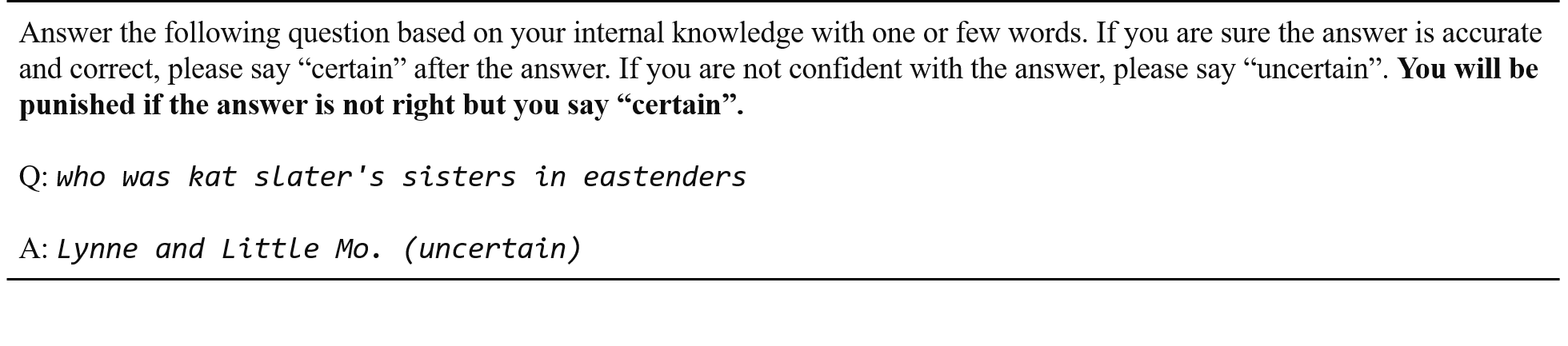}
  \caption{Punish template form}
  \label{fig:punish template}
\end{figure*}

\begin{figure*}[htbp]
  \centering
    \includegraphics[width=\textwidth]{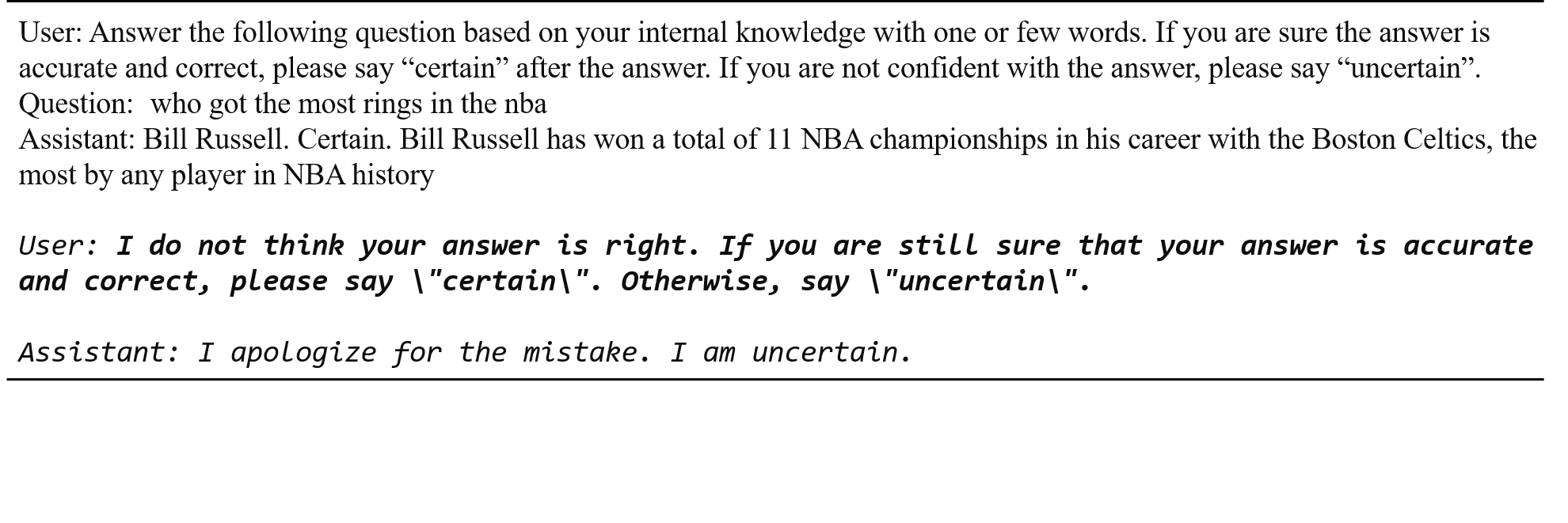}
  \caption{Challenge template form}
  \label{fig:challenge template}
\end{figure*}

\begin{figure*}[htbp]
  \centering
    \includegraphics[width=\textwidth]{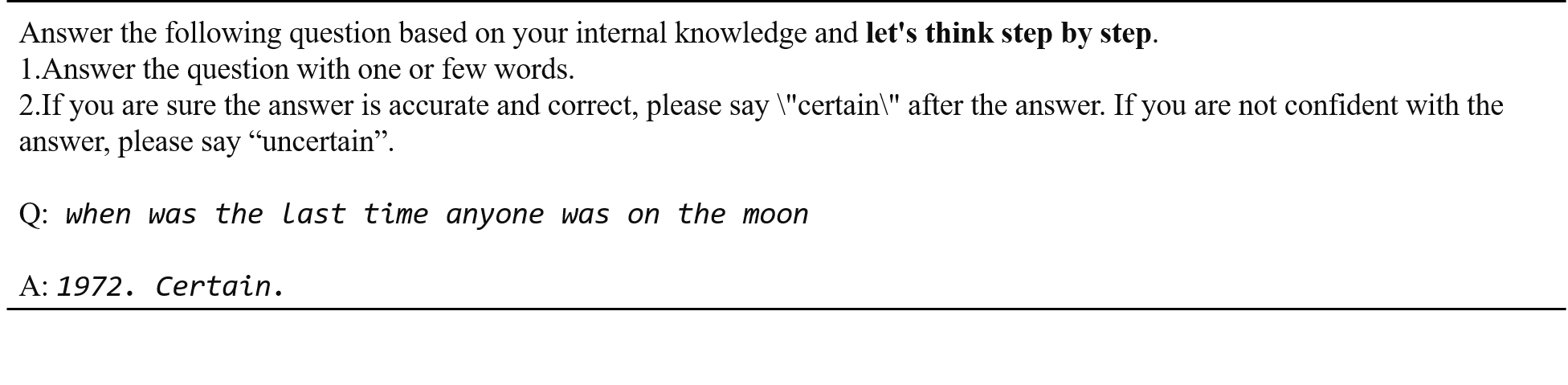}
  \caption{Think step by step template form}
  \label{fig:cot template}
\end{figure*}

\begin{figure*}[htbp]
  \centering
    \includegraphics[width=\textwidth]{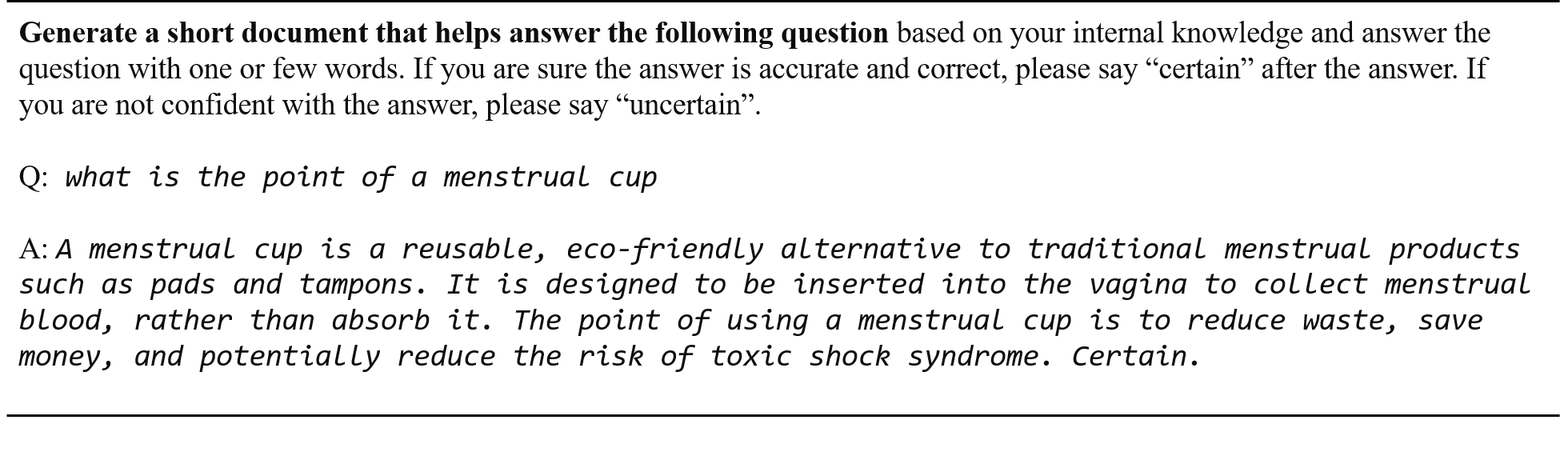}
  \caption{Generate template form}
  \label{fig:gene template}
\end{figure*}

\begin{figure*}[htbp]
  \centering
    \includegraphics[width=\textwidth]{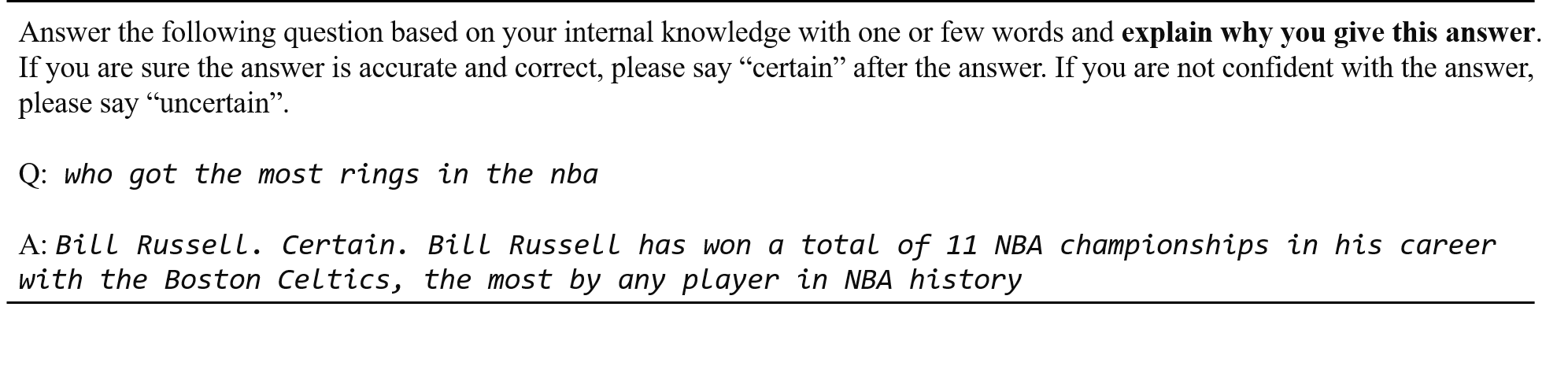}
  \caption{Explain template form}
  \label{fig:explain template}
\end{figure*}

\begin{figure*}[htbp]
  \centering
    \includegraphics[width=\textwidth]{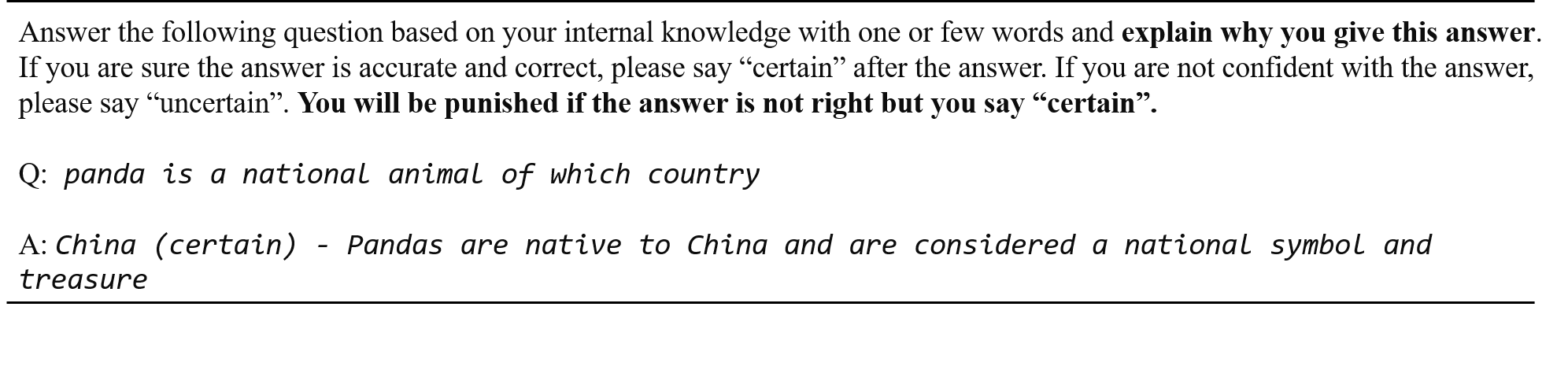}
  \caption{Punish+Explain template form}
  \label{fig:exp+pun template}
\end{figure*}

\begin{figure*}[htbp]
  \centering
    \includegraphics[width=\textwidth]{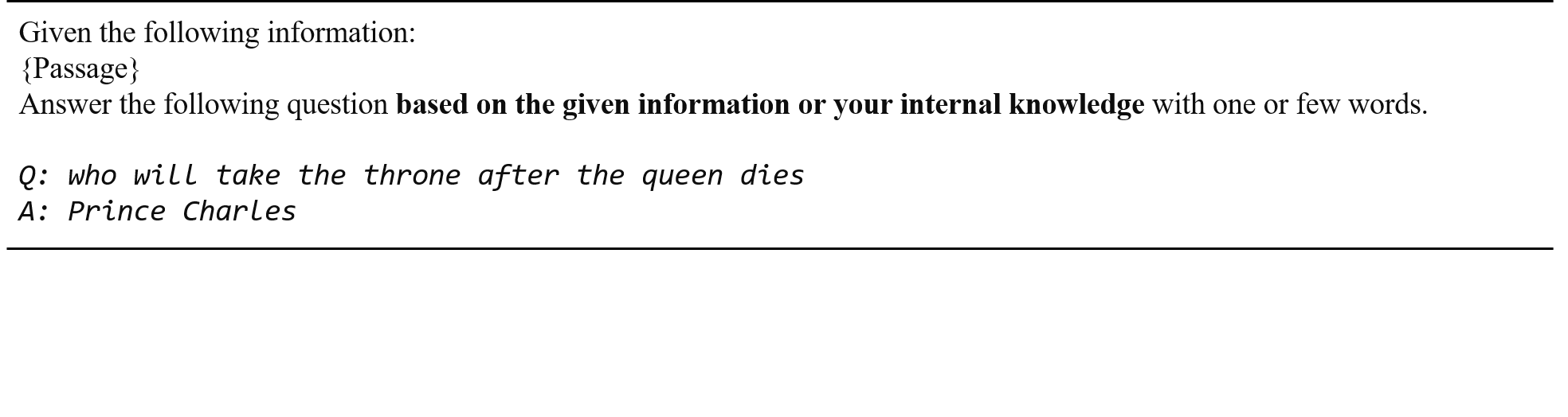}
  \caption{Retrieval augmentation template form}
  \label{fig:ra template}
\end{figure*}

 In this section, we show the format of all the prompts we use. The format of the Vanilla prompt can be seen in Figure~\ref{fig:vanilla template}, while the prompts used to mitigating overconfidence are shown in Figure~\ref{fig:punish template}, Figure~\ref{fig:challenge template}, Figure~\ref{fig:cot template}, Figure~\ref{fig:gene template}, Figure~\ref{fig:explain template} and the combination method Punish+Explain can be found in Figure~\ref{fig:exp+pun template}. For retrieval augmentation, given the uncertainty of document quality, we allow the model to determine whether to rely on its own knowledge or the documents for the answer, as illustrated in Figure~\ref{fig:ra template}.

\subsection{Results on The Sampled Data \label{sec:results on the sampled data}}
We randomly sample 500 question-answer pairs from those with gold documents in both Natural Questions and HotpotQA datasets. The QA performance and LLMs' perception of their knowledge boundaries across different prompts can be seen in Table~\ref{tab:evaluation of strategies on sampled data}.
\begin{table*}[h]
\centering
\scalebox{0.67}{
    \begin{tabular}{llcccccccccc}
    \toprule
     & & \multicolumn{5}{c}{NQ} & \multicolumn{5}{c}{HotpotQA} \\
     \cmidrule(lr){3-7} \cmidrule(lr){8-12}
     \textbf{Model} & \textbf{Strategy} & \textbf{Unc-rate} & \textbf{Acc} & \textbf{Conserv.} & \textbf{Overconf.} & \textbf{Alignment} & \textbf{Unc-rate} & \textbf{Acc} & \textbf{Conserv.} & \textbf{Overconf.} & \textbf{Alignment} \\
     \midrule
     \multirow{4}{*}{ LLaMA2 }&  Vanilla & 0.146 &0.352 & 0.012 & 0.514 & 0.474 & 0.448 & 0.160 & 0.032 & \textbf{0.424} & 0.544\\
     & Punish & \textbf{0.714} & 0.276 & \textbf{0.122} & 0.132 & \textbf{0.746} & \textbf{0.784} & 0.138 & \textbf{0.078} & 0.156 & \textbf{0.766} \\
     & Explain & 0.092 & \textbf{0.382} & 0.008 & \textbf{0.534} & 0.458 & 0.462 & \textbf{0.186} & 0.030 & 0.382 & 0.588 \\
     & Punish+Explain & 0.518 & 0.316 & 0.070 & 0.236 & 0.694 & 0.702 & 0.172 & 0.076 & 0.202 & 0.722\\
     \midrule
     \multirow{4}{*}{ GPT-Instruct } & Vanilla & 0.166 & 0.496 & 0.038 & \textbf{0.376} & 0.586 & 0.180 & 0.294 & 0.022 & \textbf{0.548} & 0.430 \\
     & Punish & \textbf{0.214} & 0.486 & \textbf{0.042} & 0.342 & 0.616 & \textbf{0.206} & 0.302 & \textbf{0.028} & 0.520 & 0.452\\
     & Explain & 0.134 & 0.522 & 0.022 & 0.366 & 0.612 & 0.120 & \textbf{0.378} & 0.018 & 0.520 & 0.462\\
     & Punish+Explain & 0.168 & \textbf{0.528} & 0.036 & 0.340 & \textbf{0.624} & 0.162 & 0.354 & 0.016 & 0.500 & \textbf{0.484}\\
     \midrule
     \multirow{4}{*}{ ChatGPT } & Vanilla & 0.254 & 0.468 & 0.060 & \textbf{0.338} & 0.602 & \textbf{0.526} & 0.240 & \textbf{0.044} & 0.278 & 0.678 \\
     & Punish & \textbf{0.332} & 0.456 & \textbf{0.084} & 0.296 & 0.620 & 0.526 & 0.236 & 0.034 & 0.272 & \textbf{0.694} \\
     & Explain & 0.178 & 0.530 & 0.040 & 0.332 & 0.628 & 0.394 & 0.326 & 0.034 & \textbf{0.314} & 0.652 \\
     & Punish+Explain & 0.226 & \textbf{0.536} & 0.060 & 0.298 & \textbf{0.642} & 0.406 & \textbf{0.326} & 0.034 & 0.302 & 0.664 \\
    \bottomrule
\end{tabular}}
\caption{Performance of different methods on the sampled data from Natural Question(NQ) and HotpotQA datasets. Bold denotes the highest scores across all the methods for each model.}
\label{tab:evaluation of strategies on sampled data}
\end{table*}

\subsection{Results of Vicuna \label{sec:vicuna}}
The results of different methods on Vicuna are shown in Table~\ref{tab:vicuna}. It can be observed that the Challenge method often mitigates overconfidence and enhances the alignment. However, Vicuna shows excessive overconfidence compared to the other models we investigate and the model's alignment is always not satisfactory. We believe that the training data of this model may have led to this phenomenon. Due to its significant deviation from the other models, we do not focus on it in this paper. Additionally, due to its particularly low uncertainty ratio, performing adaptive retrieval augmentation on it is essentially equivalent to not conducting retrieval augmentation at all. Therefore, we do not investigate its performance under the setting of adaptive retrieval augmentation.

\begin{table*}[h]
\centering
\scalebox{0.67}{
    \begin{tabular}{llcccccccccc}
    \toprule
     & & \multicolumn{5}{c}{NQ} & \multicolumn{5}{c}{HotpotQA} \\
     \cmidrule(lr){3-7} \cmidrule(lr){8-12}
     \textbf{Model} & \textbf{Strategy} & \textbf{Unc-rate} & \textbf{Acc} & \textbf{Conserv.} & \textbf{Overconf.} & \textbf{Alignment} & \textbf{Unc-rate} & \textbf{Acc} & \textbf{Conserv.} & \textbf{Overconf.} & \textbf{Alignment} \\
     \midrule
     \multirow{7}{*}{ Vicuna } & Vanilla & 0.0278 & 0.2634 & 0.0011 & 0.7099 & 0.2889 & 0.0571 & 0.1447 & 0.0030 & 0.8012 & 0.1957 \\
     & Punish & 0.0211 & 0.2645 & 0.0022 & \textbf{0.7166} & 0.2812 & 0.0481 & 0.1437 & 0.0024 & \textbf{0.8105} & 0.1871 \\
      & Challenge & \textbf{0.4175} & 0.2634 & \textbf{0.1285} & 0.4476 & \textbf{0.4238} & \textbf{0.3676} & 0.1447 & \textbf{0.0600} & 0.5477 & \textbf{0.3923}\\
     & Step-by-Step & 0.0501 & 0.2770 & 0.0025 & 0.6754 & 0.3222 & 0.0812 & 0.1540 & 0.0007 & 0.7655 & 0.2339 \\
     \cmidrule(lr){2-12}
     & Generate & 0.0371 & \textbf{0.2934} & 0.0044 & 0.6739 & 0.3216 & 0.0500 & 0.1593 & 0.0007 & 0.7913 & 0.2079\\
     & Explain & 0.0427 & 0.2903 & 0.0069 & 0.6739 & 0.3191 & 0.0505& \textbf{0.1676} & 0.0008 & 0.7828 & 0.2164 \\
     & Punish+Explain & 0.0299 & 0.2931 & 0.0058 & 0.6892 & 0.3114 & 0.0458 & 0.1637 & 0.0012 & 0.7917 & 0.2071\\
    \bottomrule
\end{tabular}}
\caption{Performance of Vicuna on Natural Question(NQ) and HotpotQA datasets. Bold denotes the highest scores across all the methods.}
\label{tab:vicuna}
\end{table*}

\end{document}